\documentclass[runningheads]{llncs}

\usepackage{eccv}

\usepackage[dvipsnames]{xcolor}

\usepackage{makecell}

\usepackage{mathtools}

\newcommand{\defeq}{\coloneqq}

\newcommand{\bI}{\mathbf{I}}

\newcommand{\bzero}{\mathbf{0}}

\newcommand{\bx}{\mathbf{x}}

\newcommand{\bmu}{{\boldsymbol{\mu}}}

\newcommand{\bSigma}{{\boldsymbol{\Sigma}}}
\usepackage{amsmath}

\usepackage{marvosym, ifsym}

\usepackage{eccvabbrv}

\usepackage{graphicx}
\usepackage{booktabs}
\usepackage{array}
\usepackage{makecell}
\usepackage{marvosym}

\usepackage[accsupp]{axessibility}

\usepackage{hyperref}

\usepackage{orcidlink}

\begin{document}

\title{GroupDiff: \\Diffusion-based Group Portrait Editing}
\titlerunning{GroupDiff: Diffusion-based Group Portrait Editing}

\author{Yuming Jiang\inst{1}\textsuperscript{*}\orcidlink{0000-0001-7653-4015}
\and
Nanxuan Zhao\inst{2}\textsuperscript{\Letter}\orcidlink{0000-0002-4007-2776}
\and
Qing Liu\inst{2}\orcidlink{0000-0003-0879-7440}
\and
Krishna Kumar Singh\inst{2}\orcidlink{0000-0002-8066-6835}
\and
Shuai Yang\inst{3}\orcidlink{0000-0002-5576-8629}
\and
Chen Change Loy\inst{1}\orcidlink{0000-0001-5345-1591}
\and
Ziwei Liu\inst{1}\orcidlink{0000-0002-4220-5958}
}

\authorrunning{Y. Jiang et al.}

\institute{College of Computing and Data Science, Nanyang Technological University
\and
Adobe Research
\and
Wangxuan Institute of Computer Technology, Peking University
\url{https://github.com/yumingj/GroupDiff}
}

\maketitle

\begin{abstract}

    Group portrait editing is highly desirable since users constantly want to add a person, delete a person, or manipulate existing persons. It is also challenging due to the intricate dynamics of human interactions and the diverse gestures. In this work, we present \textbf{GroupDiff}, a pioneering effort to tackle group photo editing with three dedicated contributions:
    \textbf{1) Data Engine:} Since there is no labeled data for group photo editing, we create a data engine to generate paired data for training. The training data engine covers the diverse needs of group portrait editing.
    \textbf{2) Appearance Preservation:} To keep the appearance consistent after editing, we inject the images of persons from the group photo into the attention modules and employ skeletons to provide intra-person guidance.
    \textbf{3) Control Flexibility:} Bounding boxes indicating the locations of each person are used to reweight the attention matrix so that the features of each person can be injected into the correct places. This inter-person guidance provides flexible manners for manipulation.
    Extensive experiments demonstrate that GroupDiff exhibits state-of-the-art performance compared to existing methods.
    GroupDiff offers controllability for editing and maintains the fidelity of the original photos.

    \keywords{Group Photo Editing \and Diffusion Models}
\end{abstract}

\makeatletter{\renewcommand*{\@makefnmark}{}
\footnotetext{$^*$This work was done when Yuming Jiang was a research intern at Adobe Research.
\newline
\textsuperscript{\Letter}Corresponding Author.
}\makeatother}

\section{Introduction}
\label{sec:intro}

Have you ever been bothered by the challenge of capturing the perfect group photo, especially during significant occasions like team-building events, family reunions, or friend gatherings? Group portrait photos become a crucial means of preserving cherished memories. Imagine this scenario, an individual must unexpectedly leave an important reunion party before the group photo is taken due to an emergency. How valuable would it be to have a tool that seamlessly adds the missing person to the photo later? Group portrait editing addresses precisely this need and more. However, as a broad topic covering various operations, group portrait editing is not an easy task with many factors that need to be considered, such as human identity, interaction, and diverse gestures. It is a difficult and tedious process even for experts with design experience and knowledge.

\begin{figure*}[t]
   \begin{center}
      \includegraphics[width=0.85\linewidth]{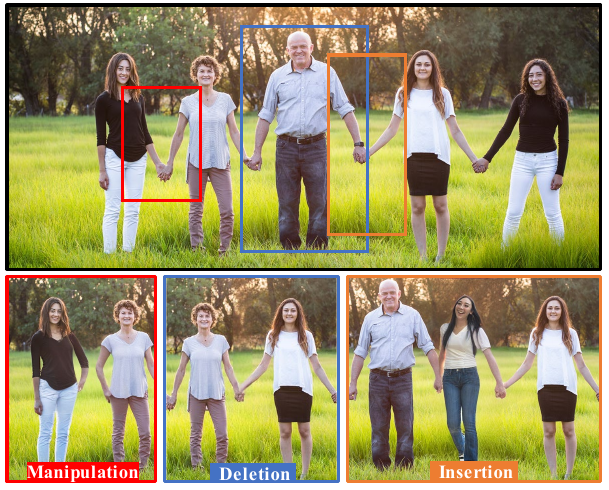}
   \end{center}
   \captionof{figure}{\textbf{Applications enabled by our GroupDiff.}
   Given a group photo, we can (a) \textbf{\color{red}{manipulate the existing person}}, (b) \textbf{\color{blue}{remove a person}}, and (c) \textbf{\color{orange}{insert a person}}.}
   \label{teaser}
\end{figure*}

However, the manipulation of group photos is challenging for two reasons: 1) the generation of the single-person image is difficult because of the complicated human structures, and 2) the generation of human interaction regions.
How to generate individual persons and their interactions naturally is an unsolved problem. 
In this work, we take the initial step to tackle this hard problem.
More specifically, we target the unique problem in group portrait editing, \ie, the synthesis of human interactions when we perform person insertion, person removal, and person manipulation, as shown in Fig.~\ref{teaser}. As personal features like facial expressions and pose orientation can be adjusted by single human editing methods~\cite{shen2020interpreting, shen2020interfacegan, ma2017pose, ma2018disentangled, liu2019neural, liu2020neural, balakrishnan2018synthesizing, men2020controllable,fu2023unitedhuman,jiang2022text2human,fu2022stylegan,jiang2023text2performer}, our work builds on the assumption that both the initial group portrait photo and the inserted person are with satisfied facial expressions and facing orientations.

To facilitate a more general setting, we aim to develop a unified framework handling three main group photo editing tasks (\ie, manipulation of existing person, person removal, and person insertion).
We propose to perform group photo editing by generating reasonable interaction regions.
Given an initial group portrait photo with masked regions indicating the intersection areas to be edited, and reference image(s) of the nearby person(s), we learn to seamlessly hallucinate the human interaction in a natural way while maintaining the identities of individuals. With the well-established generation power of diffusion models~\cite{rombach2022ldm, song2020ddim}, we choose to build our method on top of the pre-trained diffusion model~\cite{rombach2022ldm}.

However, our task still poses several unique challenges. Firstly, collecting a large set of paired training data with before-after editing is infeasible. While several works~\cite{tang2023realfill,kulal2023affordance} take the synthetic way to generate the data, they mainly work on single-person editing without considering complex human compositions and interactions. How to generate the training data fitting the natural distribution of group portrait editing is a difficult problem. Secondly, maintaining the appearance of individuals during the editing process is a formidable challenge. Users do not want the appearance to be changed, such as clothing be changed after editing. Besides, group photos often include multiple persons, and how to correctly map the reference appearance to the target person is a remaining problem. Lastly, providing users with the flexibility to specify the desired interactions according to their preferences is essential.

To this end, we propose GroupDiff, a new approach for tackling group portrait editing tasks. For the issue of lacking large-scale before-after editing data, we propose a training data generation engine that mimics the editing requirements encountered in real-world applications of group photo editing. Many factors have been considered, such as the interaction ways, the body parts' appearances, and the poses. To maintain the appearance of the persons including the inserted person and the persons in the original photo, we propose an appearance preservation diffusion model which contains person-aware modules customized on attention layers.
By taking the person(s) that the user wants to be kept as reference image(s), this model takes both inter-person and intra-person features when injecting the references into the editing process. This module can tackle the challenge posed by multiple persons through an indicator matrix, indicating the locations of each person and thus precise attention guidance.
Additionally, we take the skeleton as an intermediate representation throughout our framework design, ensuring flexibility and controllability for users.
Experiments show that our proposed GroupDiff achieves state-of-the-art performance in the task of group portrait editing.
Our contributions are summarized as follows:

\begin{itemize}
    \item We work on a challenging task of Group Portrait Editing and make the first effort to solve it in a unified framework by generating interaction regions.
    \item We propose a novel framework (\textbf{GroupDiff}) that includes person-aware modules specifically designed to maintain the appearance of reference persons, considering both inter-person and intra-person features.
    \item We introduce a simple yet effective data synthesis method to get paired data for GroupDiff training.
    \item We demonstrate the flexibility of our model on various applications such as person insertion, person removal, and interaction editing.
\end{itemize}

\section{Related Work}

\textbf{Single-Human Editing.}
There are two major directions for single-human editing: facial editing and full-body editing.
For facial editing, the target is to manipulate the facial attributes, \eg, age manipulation~\cite{yang2018learning, wang2018recurrent}, hair synthesis~\cite{olszewski2020intuitive, xing2019hairbrush} and expression manipulation~\cite{wang2018every, shen2020interpreting, shen2020interfacegan, kim2022diffusionclip,kwon2022diffusion,kwon2022diffusionit}.
As for full-body editing, it mainly includes two directions: pose transfer and virtual try-on.
The objective of pose transfer~\cite{ma2017pose, ma2018disentangled, liu2019neural, liu2020neural, balakrishnan2018synthesizing, men2020controllable} is to seamlessly transpose the visual characteristics of an individual from one pose to another.
Albahar~\etal~\cite{albahar2021pose} introduces a pose-conditioned StyleGAN framework in which the source image is warped to align with the target pose. The warping is operated according to the correspondence between the source pose and the target pose. The warped image is then employed to modulate the features.
Try-on tasks aim to transfer the clothing of the person from one image to the other image.
Traditional Try-on methods~\cite{han2018viton, chen2023size, xie2023gp, huang2022towards} mainly focus on warping the clothing to the target person.
Baldrati~\etal~\cite{baldrati2023multimodal} and Zhu~\etal~\cite{zhu2023tryondiffusion} propose to address the problem under the setting of diffusion models.
Kulal~\etal~\cite{kulal2023affordance} study the single human editing task under the human-environment interactive setting. They propose a method to insert a person into the place. 
Different from previous works, our paper focuses on group portrait editing, where the interactions and compositions among persons need to be carefully handled.

\noindent\textbf{Inpainting.}
Image inpainting is an ill-posed problem, which aims to fill in the masked regions using meaningful content. Patch-based methods~\cite{darabi2012image, hays2007scene} perform the inpainting by finding the nearest matched patch and then placing them into the holes in a proper way.
GAN-based methods~\cite{iizuka2017globally, liu2018image, liu2020rethinking, nazeri2019edgeconnect, pathak2016context} often use the high-level information extracted by the networks and then use the extracted features to guide the inpainting.
Recently, since the emergence of newer generative models, \eg diffusion models and VQGAN models~\cite{esser2021taming, oord2017neural}, researchers propose to solve the task using languages~\cite{wang2023imagen, xie2023smartbrush, ni2023nuwa}.
RealFill~\cite{tang2023realfill} learns LoRA weights~\cite{hu2022lora} to inpaint the background regions from reference images. 
However, these methods are not designed for human-specific inpainting tasks, and they do not consider the specific human priors, such as human poses, and do not have modules specifically for human appearance preservation.

\noindent\textbf{Diffusion Models.}
Diffusion models have demonstrated the capability of generating images~\cite{ho2020ddpm, sohl2015deep, song2020score}.
Recently, since the emergence of Stable Diffusion~\cite{rombach2022ldm}, it has dominated the field of image generation, especially for text-to-image generation~\cite{balaji2022ediffi, zhu2023conditional}. 
Apart from texts, ControlNet~\cite{zhang2023adding} proposes to introduce multi-modal controls, such as skeleton, sketch, depth maps etc.
Stable Diffusion is also applied to some image editing tasks~\cite{kim2022diffusionclip,kwon2022diffusion,kwon2022diffusionit,huang2023reversion}. Some editing methods~\cite{hertz2022prompt, mokady2022null} achieve image editing by modifying the cross-attention maps.
Diffusion models have been also widely applied in multiple tasks including video generation~\cite{harvey2022fdm,singer2022makeavideo,ho2022videoDM, jiang2024videobooth,wu2023freeinit}, image enhancement~\cite{saharia2022sr3,ho2022cascaded} and image composition~\cite{song2023objectstitch, chen2023anydoor, lu2023tf, zhang2023controlcom}
To take advantage of the generation power of diffusion models, our work also builds on top of this learning framework.

\section{Preliminaries}

Stable Diffusion is based on diffusion models, a kind of probabilistic model. Diffusion models generate images by iterative denoising from a noise map, which is normally sampled from the Gaussian distribution. Given a Gaussian noise map $\bx_T \sim \mathcal{N}(\bzero, \bI)$, the reverse diffusion process for sampling is expressed as follows:

\begin{align}
  p_\theta(\bx_{0:T}) &\defeq p(\bx_T)\prod_{t=1}^T p_\theta(\bx_{t-1}|\bx_t), \\
  p_\theta(\bx_{t-1}|\bx_t) &\defeq \mathcal{N}(\bx_{t-1}; \bmu_\theta(\bx_t, t), \bSigma_\theta(\bx_t, t)),
\end{align}
where $\theta$ is often parameterized using a trained network, and $t$ is the denoising timestep.

Traditional diffusion model~\cite{ho2020ddpm} operates on the pixel space. Due to the high computational cost of generating high-resolution images, the generative capability of the diffusion model is limited to low-resolution images.
Stable Diffusion~\cite{rombach2022ldm} proposes to use the VAE to project the image from pixel space to latent space. With the downscaled latent representations, the model can synthesize high-resolution images. Also, Stable Diffusion introduces texts as a condition and supports text-to-image generation.
In Stable Diffusion, $\theta$ is parameterized as a UNet with multiple self-attention modules and cross-attention modules. In each block, a self-attention module is followed by a cross-attention module. In the cross-attention module, the features $f$ are updated as follows:
\begin{align}
     f^{\prime} & = \text{softmax}(\frac{QK^T}{\sqrt{d}}) V, \\
    K &= \phi_K(f_t), V = \phi_V(f_t), Q = \phi_Q(f),
\end{align}
where $f_t$ is the text embedding, $\phi_K$, $\phi_V$, and $\phi_Q$ are the linear layers to project the features into keys, values, and queries respectively.

\section{GroupDiff}

\begin{figure}[t]
    \begin{minipage}{0.50\textwidth} \hspace{0.03\textwidth}
    \includegraphics[width=5.5cm]{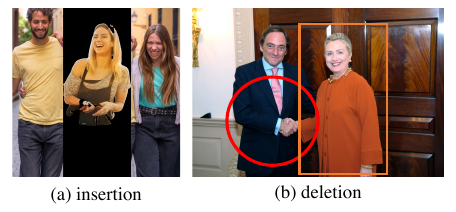}
    \end{minipage}
    \hspace{-0.04\textwidth}
    \begin{minipage}{0.50\textwidth} 
    \caption{\textbf{Illustration of Common Editing Requests.} (a) When we insert a person who only has a half-body picture, we need to adjust the interactions and make the lower part of her body complete. (b) When we are to remove a person from a group photo, we need to change the interactions and inpaint the removed region.}
    \label{illustration}
    \end{minipage}
\end{figure}

\begin{figure*}[t]
    
   \begin{center}
      \includegraphics[width=1.0\linewidth]{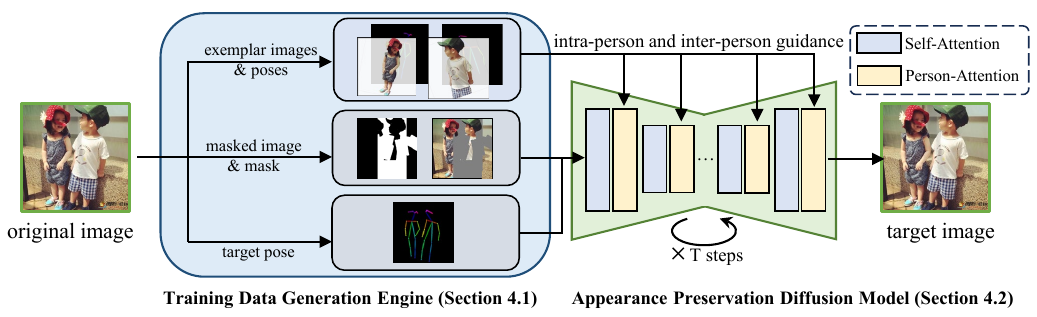}
   \end{center}
   \caption{\textbf{Overview of GroupDiff.} Starting from a group photo from the dataset, we first use the training data generation pipeline (Sec.~\ref{sec:data}) to generate paired data. Then the synthesized pair is fed into the Appearance Preservation Diffusion Model (Sec.~\ref{sec:person-aware}), where inter-person and intra-person guidance are employed to preserve the identities. 
   }
   \label{framework_overview}

\end{figure*}

In this paper, we propose a unified framework, named GroupDiff, for group portrait editing. We formulate the problem as a conditional inpainting task. As shown in Fig.~\ref{illustration}, when inserting a person, we need to inpaint the interaction regions and the missing human parts. As for removing a person, if the original person shakes their hands, we also need to inpaint the interaction region and the removed part.
To better guide the synthesis of interaction regions and offer controllability to users, we introduce the skeleton as an additional condition. For person addition, with pose controls, we can explicitly manipulate the pose of the inserted person and the neighboring person, so that their interactions are natural. As for person removal, we also need to change the pose of the existing person to make it to be natural, if there are some human interactions between the removed person and existing persons.
The overview of our proposed GroupDiff is shown in Fig.~\ref{framework_overview}. 
We tackle the problem of group photo editing from two perspectives.
From the data side, we propose a comprehensive training data generation engine to synthesize paired data, which will be introduced in Sec.~\ref{sec:data}.
From the model side, we take the stable diffusion as the backbone~\cite{rombach2022ldm}.
Except for the noisy map, our model also takes an input image masked with the region to be filled, the corresponding mask, and the target skeleton map as inputs.
These inputs are concatenated together before feeding them to the network. The masked region is filled with gray color.
The skeleton map offers the user an option to specify the desired human interaction. 
As the masked region often covers the area that should remain the same after editing, such as the clothing and skin color, our model also takes the reference images of persons as conditions. Each reference person is given as a separate image. The reference images can be the inserted person or nearby persons around the interaction regions. We inject the reference images through intra-person and inter-person guidance to let the model learn to keep the appearance details after editing.

\subsection{Training Data Generation Engine}
\label{sec:data}

The training of diffusion-based models requires large-scale paired data. However, there is no such dataset for group photo editing. Given the difficulties of conducting group portrait photo editing, collecting a large amount of before-after editing data is infeasible. Therefore, finding a way to generate pairwise data to support the training is demanded.
Since we pose our task as an inpainting task, we start from the existing group photo dataset~\cite{zhao2018understanding, li2017multiple, nie2017generative}.
Considering the real use cases, we design a comprehensive training data generation engine.

\begin{figure}[t]
    \begin{minipage}{0.70\textwidth} \hspace{0.03\textwidth}
    \includegraphics[width=7.5cm]{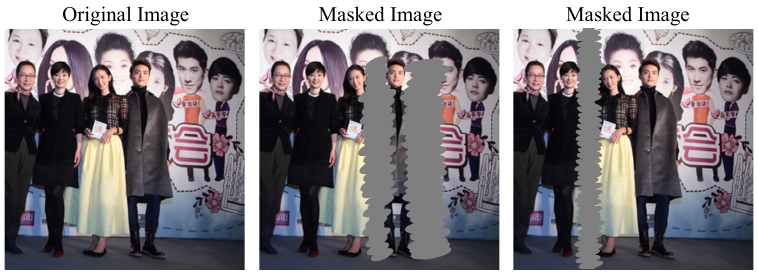}
    \end{minipage}
    \hspace{-0.04\textwidth}
    \begin{minipage}{0.30\textwidth} 
    \caption{\textbf{Coarse Level Training Data Generation for Person Interaction.} At the coarse level, we generate masks according to the bounding boxes of persons.}
    \label{data_generation_a}
    \end{minipage}
\end{figure}

\begin{figure}[t]
    \begin{minipage}{0.70\textwidth} \hspace{0.03\textwidth}
    \includegraphics[width=7.5cm]{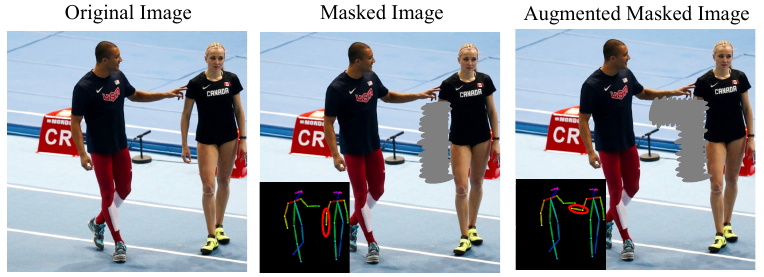}
    \end{minipage}
    \hspace{-0.04\textwidth}
    \begin{minipage}{0.30\textwidth} 
    \caption{\textbf{Fine Level Training Data Generation for Person Interaction.} At the fine level, we generate masks according to the skeleton and augmented skeleton.}
    \label{data_generation_b}
    \end{minipage}
\end{figure}

\noindent\textbf{Person Interaction.}
Person interactions encompass a diverse set of gestures that form the essence of social dynamics including physical interactions, such as hugging, shaking hands, and raising arms, and many others. These free-form interactions increase difficulties in approximating the real distribution. Besides, different editing operations may incur different challenges. For example, when adding a person into the crowd, the surrounding context also should be completed, such as the area above the person's head.
To enforce our model to better generate diverse interactions under different conditions, we use a hierarchical way for synthesizing paired data by randomly masking corresponding regions in group photo images. The coarser scale targets scenarios of large modification such as person insertion and removal, which requires inpainting a large region(Fig.~\ref{data_generation_a}). The finer scale targets scenarios of small modification such as interaction manipulation, which needs more fine-grained controls of masked regions(Fig.~\ref{data_generation_b}).

On a coarser scale, we first identify different persons through labeled bounding boxes. We assume that the interaction regions are often near the boundary of each bonding box. During training, we randomly select single or multiple bounding boxes and mask out the region near the boundary. 
Suppose the coordinate of the top-left point of the bounding box is $(x_1, y_1)$ and the right-bottom point is $(x_2, y_2)$. Then the width of the bounding box is $w = x_2 - x_1$. Then the left boundary region of the bounding box can be denoted as the rectangle with $(x_1 - r \cdot w, y_1)$ as the top-left corner and $(x_1 + r \cdot w, y_2)$ as the right-bottom corner. The right boundary of the bounding box can be obtained in a similar way. We simply treat the boundary regions as the interaction regions. When synthesizing data, we will randomly mask the left boundary, right boundary, or both boundaries. Empirically, we randomly sample $r$ from $\left [  0.1, 0.2 \right ]$.
As shown in the second image in Fig.~\ref{data_generation_a}, we mask both boundary regions.
Sometimes, we extend the mask outside the bounding box to cover the whole column of the boundary region as shown in the third image of Fig.~\ref{data_generation_a}. That is to set the top-left coordinate as $(x_1 - r \cdot w, 0)$ and the right-bottom coordinate as $(x_1 + r \cdot w, h)$, where $h$ is the height of the image.
To avoid facial regions being masked, we use human parsing to unmask the facial regions.
To make the shape of the mask closer to the users' inputs, we also augment the shape of the mask into a brush-like one.

On a finer scale, we introduce the skeleton to control the person interaction, especially for the hand and arm positions. Involving the skeleton helps users to better specify the desired human interaction. The correctness of the skeleton is important to ensure the data generation quality, we use the state-of-the-art skeleton detection method~\cite{mmpose2020} for extraction. A naive way of generating the mask is to segment out the arm and hand regions with a tight bounding box. However, this simple solution has a critical problem as the mask shape itself will leak the skeleton information. For example, in Fig.~\ref{data_generation_b}, if we strictly follow the skeleton, we can obtain the masked image as shown in the second column. In this case, the model may overlook the skeleton condition during training because the masked image itself has leaked the skeleton information. As a result, the obtained model fails to generate images conditioned on skeletons at inference time. Therefore, we adopt data augmentations to avoid this issue. Except for the tight bounding box mask around arms and hands, we randomly rotate the arms or hands in a different direction. As shown in the third column of Fig.~\ref{data_generation_b}, we randomly rotate the right arm and obtain the augmented skeleton. We mask both arm regions according to the original skeleton and augmented skeleton.

\begin{figure}[t]
    \begin{minipage}{0.70\textwidth} \hspace{0.03\textwidth}
    \includegraphics[width=7.5cm]{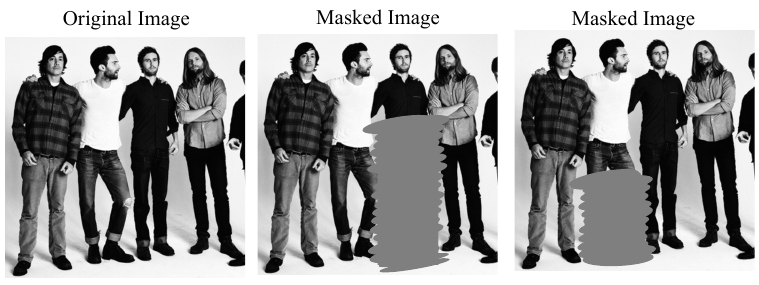}
    \end{minipage}
    \hspace{-0.04\textwidth}
    \begin{minipage}{0.30\textwidth} 
    \caption{\textbf{Training Data Generation for Body Completion.} We randomly mask the lower parts of the body so that the model has the capability of body completion.}
    \label{data_generation_c}
    \end{minipage}
\end{figure}

\noindent\textbf{Body completion.}
In certain scenarios, our model also needs to complete the body part(s) of a person. For example, when there are occlusions among different persons, removing one person needs the body completion of the surrounding people. When inserting a person, if the reference image only contains a partial body, in this case, the model needs to learn to complete the rest parts. 
The data synthesis for the body completion is straightforward. We randomly mask out the lower body parts. Still, $(x_1, y_1)$ represents the top-left point of the bounding box and $(x_2, y_2)$ represents the right-bottom point. We randomly mask the region from $(x_1, y_1 + r \cdot (y_2 - y_1)$ to $(x_2, y_2)$, where $r$ is sampled from $\left [ 0.5, 0.9 \right ]$. 
Some examples are shown in Fig.~\ref{data_generation_c}.

\begin{figure*}[t]
   \begin{center}
      \includegraphics[width=0.9\linewidth]{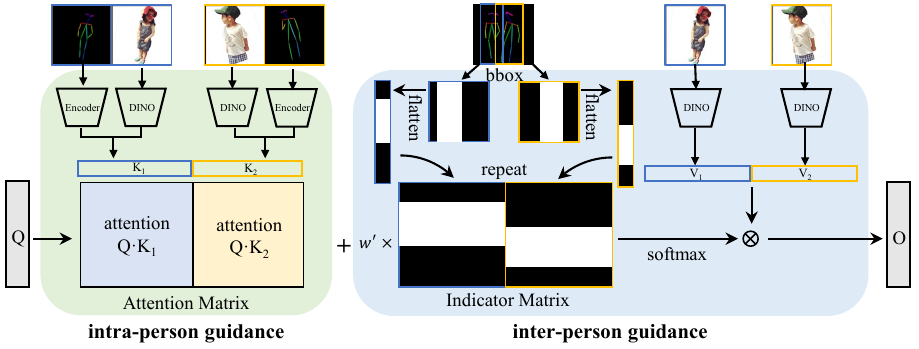}
   \end{center}
   \caption{\textbf{Overview of Person-Aware Attention Module for Appearance Preservation.} We employ intra-person guidance and inter-person guidance to better preserve the appearance. The intra-person guidance concatenates the pose features and image features as keys.
  The inter-person guidance is provided by using the information deduced from the positions of the persons. The indicator matrix is employed to assign higher weights to corresponding values in the attention matrix.
  }
   \label{person_aware_framework}
\end{figure*}

\subsection{Person-Aware Attention for Appearance Preservation}
\label{sec:person-aware}

Our model needs to preserve the appearance of persons after editing. Since the content cannot be revealed if it is covered with masks during inference, we take the persons that users want to maintain the appearance out as reference images. To ensure details are encoded into the model accurately, inspired by previous works~\cite{chen2023anydoor,kulal2023affordance}, we choose to inject reference image(s) through the cross-attention module. 
The reference images during the training are obtained as follows: 1) segment the person(s), and 2) fill the background with white.
However, feeding all reference images through concatenation to the model like previous methods does not work in our case. This can cause a mismatch problem between the reference and target persons because the attention matrix is applied among query features and features of all reference images. The model will confuse which part to attend if multiple reference images are provided. We thus introduce a person-aware cross-attention mechanism for preserving content details. 

Instead of sending raw pixels to the cross-attention module, we first encode each reference image $I_i$ using DINOv2~\cite{oquab2023dinov2} into a feature embedding $f_{I_i} \in \mathbb{R}^{256 \times 768}$, which has been validated to be effective for image generation and editing tasks~\cite{chen2023anydoor}. If we follow the conventional cross-attention operation, after concatenating features together as $F=[f_{I_1}, f_{I_2}, ..., f_{I_N}]$, we obtain the value embedding $V \in \mathbb{R}^{(256 \cdot N) \times d}$ 
and key embedding $K \in \mathbb{R}^{(256 \cdot N) \times d}$ through linear layers individually to compute the output feature $O$ as:
\begin{align}
    & O = \text{softmax}(\frac{QK^T}{\sqrt{d}})V, \\
    & K = \phi_K(F), V = \phi_V(F),
\end{align}
where $d$ denotes feature dimension, and the query embedding $Q \in \mathbb{R}^{HW \times d}$ comes from the output of the last layer. $H$ and $W$ refer to the spatial resolution of features from the last layer (a.k.a self-attention layer).
Based on this formulation, we introduce intra-person and inter-person guidances to solve our problem. 

\noindent\textbf{Intra-Person Guidance.} For the intra-person guidance, it is essential to let the model know the correspondences between pixel appearances with body components for each reference image. Then the model can adaptively learn the mapping to restore accurate appearances. We introduce the skeleton pose $P_{I_i}$ of exemplar images $I_i$ as the intra-person indicators as shown in Fig.~\ref{person_aware_framework}. Similarly, we first extract pose features $p_{I_i}$ through a separately learned encoder, and compute the key embedding with an MLP after concatenating the image and pose features:
\begin{align}
    & \hat{K} = \phi_K([K_1, K_2, ..., K_N]), \\
    & K_n = \text{MLP}([f_{I_n}, p_{I_n}]),
    % & F_{I+P} = [[f_{I_1}, p_{I_1}], [f_{I_2}, p_{I_2}], ..., [f_{I_N}, p_{I_N}]], 
\end{align} 
where $[\cdot]$ is the concatenation operation and $\text{MLP}(\cdot)$ is the layers to process the concatenated features.
The attention matrix is obtained as follows:
\begin{equation}
M_{attn} = Q\hat{K}^T \in \mathbb{R}^{H W \times 256 \cdot N}.
\end{equation}

It should be noted that we only introduce the pose information into keys and do not embed the pose information into values. The motivation lies in that pose information is an indicator of the body component and it should not be fused with the appearance information in keys.

\noindent\textbf{Inter-Person Guidance.} For inter-person guidance, as shown in Fig.~\ref{person_aware_framework}, we introduce the indicator mask to specify the locations of each reference person. The indicator masks $\{m_i \in \mathbb{R}^{H \times W}\}$ are obtained by bounding boxes around each person. The bounding boxes cover the full body.
For each reference image, we reshape the corresponding indicator mask $m_i$ into a flattened tensor (whose shape is $\mathbb{R}^{HW}$) and repeat it into a matrix so that it has the shape of $\mathbb{R}^{HW \times 256}$.
The indicator matrix $M_{ind} \in \mathbb{R}^{H W \times 256 \cdot N}$ is obtained by concatenating the matrices of all reference images. 
We add the indicator matrix to the attention matrix before applying the softmax operation. With this operation, when calculating the final output, the corresponding location will have higher weights to attend to the correct reference features. %from their corresponding exemplar embeddings.
The operation is computed as:
\begin{equation}
O = softmax(\frac{M_{attn} + w^{\prime} \cdot M_{ind}}{\sqrt{d}}) V,
\end{equation}
where the value embedding $V$ is obtained from the DINO feature of reference images without involving pose information.
In practice~\cite{balaji2022ediffi}, we multiply the indicator matrix by a scale factor $w^{\prime}$, which is obtained as follows:
\begin{equation}
    w^{\prime} = w \cdot log(1 + \sigma) \cdot max(M_{attn}),
\end{equation}
where $w$ is the user-specified factor, and $\sigma$ corresponds to the noise level at different diffusion steps.

\section{Experiments}

\subsection{Implementation Details}
Our method is implemented using PyTorch. The model is trained using eight NVIDIA A100 GPUs. We use the Adam optimizer to optimize the model. The beta is set as the default value. We set the learning rate as $10^{-4}$. The learning rate schedular is LambdaLinearScheduler. The number of warm-up steps is set as 2500. 
The batch size is set as 4 per GPU, and the global batch size is 32.
The model is initialized with the Stable-Diffusion v1.5 Inpainting model.
The original Stable Diffusion Inpainting model has 9 channels as input. We concatenate the skeleton as an additional input, and thus the number of channels is 12. For the first 9 channels of the first layer, we inherit the weights of Stable Diffusion Inpainting model and initialize the remaining three channels with zero weights.
The model is trained for 80 epochs. The resolution of the image we use for training is $512 \times 512$. 
We use the LV-MHP-v2 dataset~\cite{zhao2018understanding, li2017multiple, nie2017generative}, which is composed of $25,403$ group photo images. With each image, there are human parsing annotations. For human parsing, we directly use the annotations provided in the dataset. We use MMPose~\cite{mmpose2020} to predict the skeleton. When editing group photos at inference time, we adjust the lighting using the off-the-shelf model~\cite{sofiiuk2020harmonization}.

\subsection{Comparison Methods}
\noindent\textbf{SD Inpainting~\cite{rombach2022ldm}.} Stable Diffusion has provided the pretrained weights for inpainting. Here, we use the Stable Diffusion Inpainting (SD Inpainting) model as a baseline method. For person insertion, the interaction regions and the regions surrounding the inserted person are masked. For person removal, we mask the regions of the person to be removed. At inference time, SD Inpainting model is fed with the same masked image and masks as ours.
Apart from pretrained SD Inpainting model, we also compare our model with two more baselines, which are adapted from the pretrained SD Inpainting model. 
The first baseline is the finetuned SD Inpainting model. We use our synthesized data to finetune the SD Inpainting model. 
The second baseline is the finetuned SD Inpainting model with skeleton control. Similar to our method, we add skeleton control to the SD inpainting model. We concatenate the skeleton with other inputs. Then we finetune the model using our generated data pair. 

\noindent\textbf{Paint by Example~\cite{yang2022paint}.} Paint by Example is an exemplar-based image editing method. It fills masked regions using exemplar images. The exemplar image is fed into an image encoder, and then the extracted features are fed into the cross-attention module. 
For person insertion, we feed the model the full mask covering the place where the person is to be inserted and the single-person image as the exemplar image. 
For person deletion, the mask region is the area where the person is removed, and the exemplar image is a crop of background.

\begin{figure*}[t]
   \begin{center}
      \includegraphics[width=1.0\linewidth]{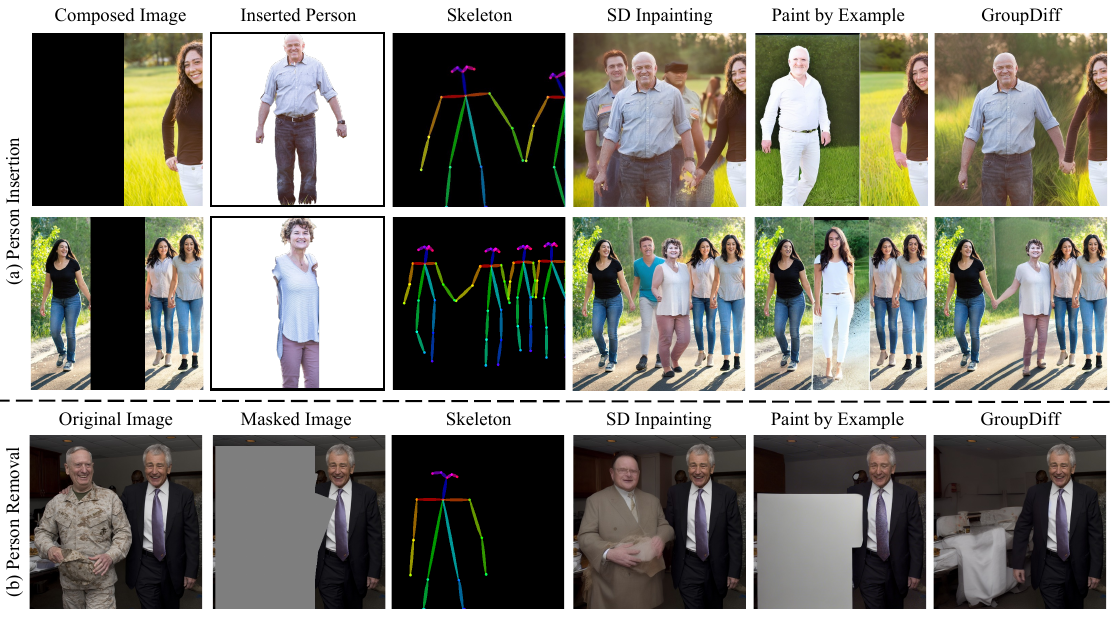}
   \end{center}
   \caption{\textbf{Qualitative Comparisons.} We compare our proposed GroupDiff against SD Inpainting~\cite{rombach2022ldm} and Paint by Example~\cite{yang2022paint}. (a) We show two examples of person insertion. (b) We show one example of person removal. 
   }
   \label{quali}
\end{figure*}

\subsection{Qualitative Comparisons}
Figure~\ref{quali} shows some qualitative comparisons of our method. 
The results of person insertion are shown in Fig.~\ref{quali}(a).
In the first row, we aim to add the old man to the image and make their hands hold together. 
According to the original image and the skeleton, we can get the masked image covering the interaction regions and the background regions. 
For the SD Inpainting model, the model adds a lot of persons behind the old man. Also, due to the lack of skeleton control, the interaction regions are not filled as expected.
For Paint by Example, it inserts the old man into the left side of the photo and the obtained image looks to be composed of two split parts. Also, the appearance of the inserted person changed a lot.
As for our proposed GroupDiff, it successfully inserts the person harmoniously. The hands of the persons correctly follow the given skeleton mask.

In the second example of Fig.~\ref{quali}(a), a lady is to be inserted into the middle of the image. After insertion, the hands of the lady are supposed to be connected with neighboring persons as indicated by the skeleton mask. Still, the SD Inpainting model tends to add another person into the middle. The interaction regions are not synthesized as expected. For the Paint by Example, a lady is inserted into the photo, however, the identity of the person is not the one indicated in the exemplar image. 
By contrast, our model successfully inserts the target person into the photo with reasonable interactions.

In Fig.~\ref{quali}(b), we show one example of person removal. We aim to remove the person standing on the left. The mask is obtained according to the bounding box of the person to be removed. After removing the person, we need to adjust the postures of the remaining person. Otherwise, the pose of the remaining person would look weird. 
Using the SD Inpainting model, there is another person present in the final image. For the Paint by Example model, we feed a part of background as the exemplar image.
The masked region is filled with some random colors. 
The final column shows our result. By removing the person and feeding the adjusted skeleton, our model can synthesize a natural photo.

\subsection{Quantitative Comparisons}
As collecting large well-curated test sets is difficult, we conduct the quantitative comparisons on a synthetic dataset that the training has never seen. We use the validation set of LV-MHP-v2~\cite{li2017multiple}, which has no overlap with our synthesized training data. We report the PSNR, SSIM, FID, CLIP-I, and DINO scores in Table~\ref{tab:quant}. Compared with SD Inpainting and Paint by Example, our proposed GroupDiff has significantly better performance.
Table~\ref{tab:quant} shows that the quantitative results improve after we fine-tune the pretrained SD inpainting model.
Besides, skeleton control helps to improve the performance. The PSNR and SSIM values increase, while the FID score decreases.
However, even with skeleton control, there is still a performance gap between this model and our proposed GroupDiff, which indicates the necessity of our proposed modules.

\begin{table}[t]
  \centering
    \caption{\textbf{Quantitative Comparisons.} We report PSNR, SSIM, FID, CLIP-I and DINO scores on synthesized data. Compared to baseline methods, our method achieves significantly better performance. ``Random Mask'' and ``Mask One Person'' refer to two ablation studies on the training data generation pipeline. ``w/o inter-person'' and ``w/o intra-person'' are two ablation studies on the person-aware module.}
    \scriptsize{
    \begin{tabular}{l|c|c|c|c|c}
    \Xhline{1pt}
    \textbf{Methods} & \textbf{PSNR $\uparrow$} & \textbf{SSIM $\uparrow$} & \textbf{FID $\downarrow$} & \textbf{\makecell{CLIP-I $\uparrow$}} & \textbf{\makecell{DINO $\uparrow$}}\\ \Xhline{1pt}
    SD-Inpainting~\cite{rombach2022ldm} & 17.4148  & 0.7562 & 24.1276 & 79.5467 & 91.2230 \\ 
    Paint-by-Example~\cite{yang2022paint}  & 14.4848  & 0.6298 & 34.8999 & 74.1095  & 87.4915 \\ 
    Finetuned SD Inpainting & 18.3132  & 0.7846 & 20.4833 & 82.4550 & 93.1132 \\
    Finetuned SD Inpainting with Skeleton & \underline{18.5688}  & \underline{0.7868} & \underline{20.3351} & \underline{83.3211} & \underline{93.5158} \\
    \hline \hline
    GroupDiff (Ours)  & \textbf{19.4569}  & \textbf{0.8033} & \textbf{18.1035} & \textbf{86.8967} & \textbf{95.0284}\\ 
    \Xhline{1pt}
  \end{tabular}
  }
  \label{tab:quant}
\end{table}

\subsection{Ablation Studies}
We report quantitative results for ablation studies in Table~\ref{tab:quant_ablation}. 
We will show visual comparisons in supplementary files.

\noindent\textbf{Training Data Generation Engine.}
We first investigate the performance of the model with ``Random Mask''.
The ablation model is trained using the randomly masked group photo images. The images are randomly masked regardless of the bounding boxes of the persons and the skeletons. Quantitative results reported in Table~\ref{tab:quant_ablation} demonstrate that 
the model trained with the random mask has inferior performance. As discussed before, with the random mask, the model can infer the pose from the masked image and thus the capability of conditioning on the skeleton is limited. 
We also experiment with the ``Mask One Person'' strategy, we synthesize the data by directly masking one person according to the bounding box~\cite{kulal2023affordance}. Still, by randomly masking a single person, the model has limited capability of inpainting interaction regions. As shown in Table~\ref{tab:quant_ablation}, our model has a significantly better performance. 

\begin{table}[t]
  \centering
    \caption{\textbf{Quantitative Comparisons for Ablation Studies.} We conduct ablation studies from two persepctives. ``Random Mask'' and ``Mask One Person'' refer to two ablation studies on the training data generation pipeline. ``w/o inter-person'' and ``w/o intra-person'' are two ablation studies on the person-aware module.}
    \footnotesize{
    \begin{tabular}{l|c|c|c}
    \Xhline{1pt}
    \textbf{Methods} & \textbf{PSNR $\uparrow$} & \textbf{SSIM $\uparrow$} & \textbf{FID $\downarrow$} \\ \Xhline{1pt}
    Random Mask & 17.3299  & 0.7727 & 26.2193 \\
    Mask One Person & 17.9515  & 0.7777 & 22.7869 \\ \hline \hline
    w/o inter-person & 19.2612  & 0.7997 & 18.4946 \\
    w/o intra-person & \underline{19.4236}  & \textbf{0.8034} & \underline{18.2274} \\ \hline \hline
    GroupDiff (Ours)  & \textbf{19.4569}  & \underline{0.8033} & \textbf{18.1035} \\ 
    \Xhline{1pt}
  \end{tabular}
  }
  \label{tab:quant_ablation}
\end{table}

\noindent\textbf{Person-aware Attention.}
As shown in Table~\ref{tab:quant_ablation}, after removing the inter-person guidance, the model will have inferior PSNR, SSIM, and FID scores, which means the performance of the model ``w/o inter-person'' is inferior.
As for removing the intra-person guidance, the model ``w/o intra-person'' has a better SSIM score but inferior PSNR and FID performance. We show visual examples in the supplementary file to demonstrate the effectiveness of this module.

\section{Conclusion}

In this work, we introduce a pioneering solution to the intricate challenging problem of group portrait editing. We formulate the task as an inpainting problem, addressing issues such as limited labeled data, appearance preservation, and manipulation flexibility. Our approach includes a data generation pipeline mimicking the real editing scenarios and a person-aware appearance preservation module for consistent editing results. 
Extensive experiments demonstrate the superiority of our proposed method.
With the well-known difficulties of human generation and manipulation even for a single person, there is still ample room for refinement and innovation in our task. 
We hope our work can inspire the exploration of this important task.

\noindent\textbf{Limitations.} Our proposed GroupDiff makes an assumption that the inserted person image has correct facial expression and facing orientations. If a person image from the side view is given, our method cannot insert the person into the group photo where all people are facing front. Besides, in the current model, the lighting of the inserted person is changed as a preprocessing step. Thus, the ability to adjust the lighting is capped by the capability of the pretrained model. More powerful models can be used to better adjust the lighting~\cite{guerreiro2023pct, jiang2021ssh, ke2022harmonizer}.

\noindent\textbf{Potential Negative Impact.} The proposed method can be used to generate fake images by putting a lot of people who don't know each other in a group photo. It may be negatively used to fabricate certain facts.

\noindent\textbf{Acknowledgement.} This study is supported by the Ministry of Education, Singapore, under its MOE AcRF Tier 2 (MOET2EP20221- 0012), and NTU NAP.
% , and under the RIE2020 Industry Alignment Fund – Industry Collaboration Projects (IAF-ICP) Funding Initiative, as well as cash and in-kind contribution from the industry partner(s).

\bibliographystyle{splncs04}
\bibliography{main}

% \documentclass[runningheads]{llncs}

% ---------------------------------------------------------------
% Include basic ECCV package

% \usepackage{eccv}

% OPTIONAL: Un-comment the following line for a version which is easier to read
% on small portrait-orientation screens (e.g., mobile phones, or beside other windows)
% \usepackage[mobile]{eccv}

% \input{preamble}
% % ---------------------------------------------------------------
% % Other packages

% % Commonly used abbreviations (\eg, \ie, \etc, \cf, \etal, etc.)
% \usepackage{eccvabbrv}

% % Include other packages here, before hyperref.
% \usepackage{graphicx}
% \usepackage{booktabs}
% \usepackage{array}
% \usepackage{makecell}

% % The "axessiblity" package can be found at: https://ctan.org/pkg/axessibility?lang=en
% \usepackage[accsupp]{axessibility}  % Improves PDF readability for those with disabilities.

% ---------------------------------------------------------------
% Hyperref package

% It is strongly recommended to use hyperref, especially for the review version.
% Please disable hyperref *only* if you encounter grave issues.
% hyperref with option pagebackref eases the reviewers' job, but should be disabled for the final version.
%
% If you comment hyperref and then uncomment it, you should delete
% main.aux before re-running LaTeX.
% (Or just hit 'q' on the first LaTeX run, let it finish, and you
%  should be clear).

% \usepackage{hyperref}
% \appendix

% Support for ORCID icon
% \usepackage{orcidlink}
\renewcommand\thesection{\Alph{section}}
\renewcommand\thefigure{A\arabic{figure}}
\renewcommand\thetable{A\arabic{table}}

% \begin{document}

\title{GroupDiff: \\Diffusion-based Group Portrait Editing
\\
Supplementary Files} 

\titlerunning{GroupDiff: Diffusion-based Group Portrait Editing}

\author{Yuming Jiang\inst{1}\textsuperscript{*}\orcidlink{0000-0001-7653-4015} 
\and
Nanxuan Zhao\inst{2}\textsuperscript{\Letter}\orcidlink{0000-0002-4007-2776} 
\and
Qing Liu\inst{2}\orcidlink{0000-0003-0879-7440}
\and
Krishna Kumar Singh\inst{2}\orcidlink{0000-0002-8066-6835}
\and
Shuai Yang\inst{3}\orcidlink{0000-0002-5576-8629}
\and
Chen Change Loy\inst{1}\orcidlink{0000-0001-5345-1591}
\and
Ziwei Liu\inst{1}\orcidlink{0000-0002-4220-5958}
}

\authorrunning{Y.~Jiang et al.}

\institute{
College of Computing and Data Science, Nanyang Technological University 
\and
Adobe Research 
\and
Wangxuan Institute of Computer Technology, Peking University
}

\maketitle

\section{Visual comparisons for ablation study}
\begin{figure*}
   \begin{center}
      \includegraphics[width=0.80\linewidth]{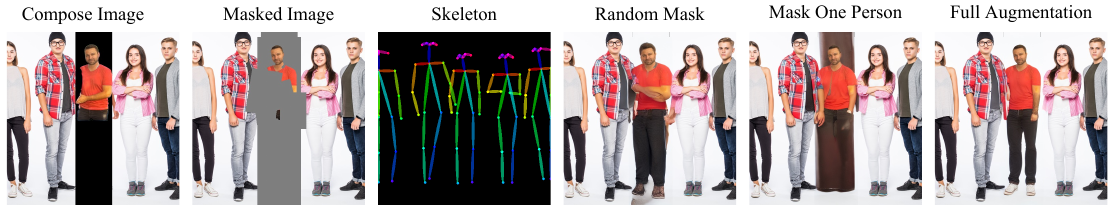}
   \end{center}
  \vspace{-20pt}
   \caption{\textbf{Ablation Study on Training Data Generation.}
   }
    % \vspace{-10pt}
   \label{ablation_data}
\end{figure*}

\noindent\textbf{Training Data Generation.}
The result of ``Random Mask'' is shown in the fourth column of Fig.~\ref{ablation_data}. The model trained with the random mask is not able to repose the inserted person. Our model can successfully repose the inserted person as indicated in the skeleton map.
The result of ``Mask One Person'' strategy is shown in Fig.~\ref{ablation_data}. The model has the capability of reposing the inserted person. However, the model fails to complete the lower body of the inserted person as the model is always fed with the exemplar image having the same body components as the target image. By contrast, our model can successfully complete the lower part with reasonable content.

\begin{figure*}
   \begin{center}
      \includegraphics[width=0.80\linewidth]{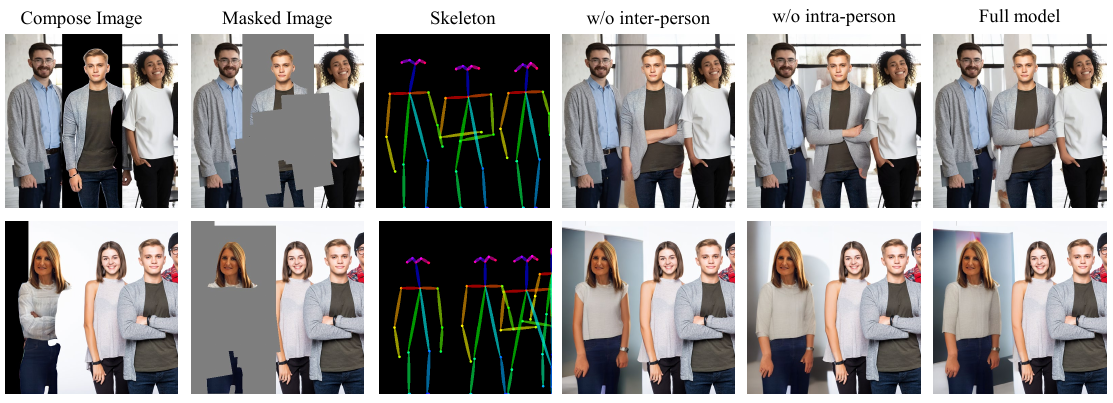}
   \end{center}
  \vspace{-20pt}
   \caption{\textbf{Ablation Study on Person-aware Appearance Preservation Module.}
   }
    % \vspace{-20pt}
   \label{ablation_attn}
\end{figure*}

\noindent\textbf{Person-aware Appearance Preservation:} Without ``inter-person guidance'' (\ie, without the whole person-aware module), the results are shown in the fourth column of Fig.~\ref{ablation_attn}. In the first example, without the whole person-aware module, the model cannot successfully condition the skeleton pose. The right hand of the man is not in the correct place. In the second example, without the person-aware module, the model confuses the appearance of the neighboring person, and the lady on the left wears short-sleeve upper clothing, which is not consistent with the original exemplar image. However, with inter-person guidance (result in the fifth column and the last column), the model can successfully transfer the correct clothing from the exemplar image.
The intra-person guidance is introduced to enhance the capability of the model to change the pose of the inserted person. As indicated in the first example in Fig.~\ref{ablation_attn}, without intra-person guidance, the model cannot successfully repose the left hand of the man. After introducing the intra-person guidance, the model can place the hands of the inserted person in the correct place.

\section{More Qualitative Results}
\begin{figure*}
   \begin{center}
      \includegraphics[width=0.80\linewidth]{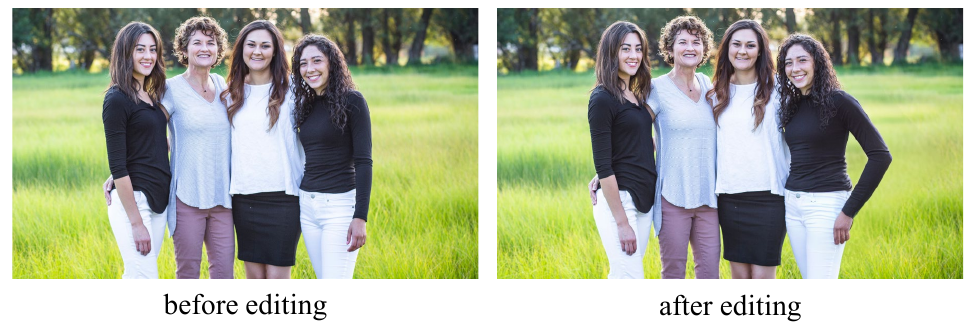}
   \end{center}
  \vspace{-20pt}
   \caption{\textbf{More Visual Results on Editing Existing Person.}
   }
   % \vspace{-20pt}
   \label{Quali_editing}
\end{figure*}

\begin{figure*}
   \begin{center}
      \includegraphics[width=0.80\linewidth]{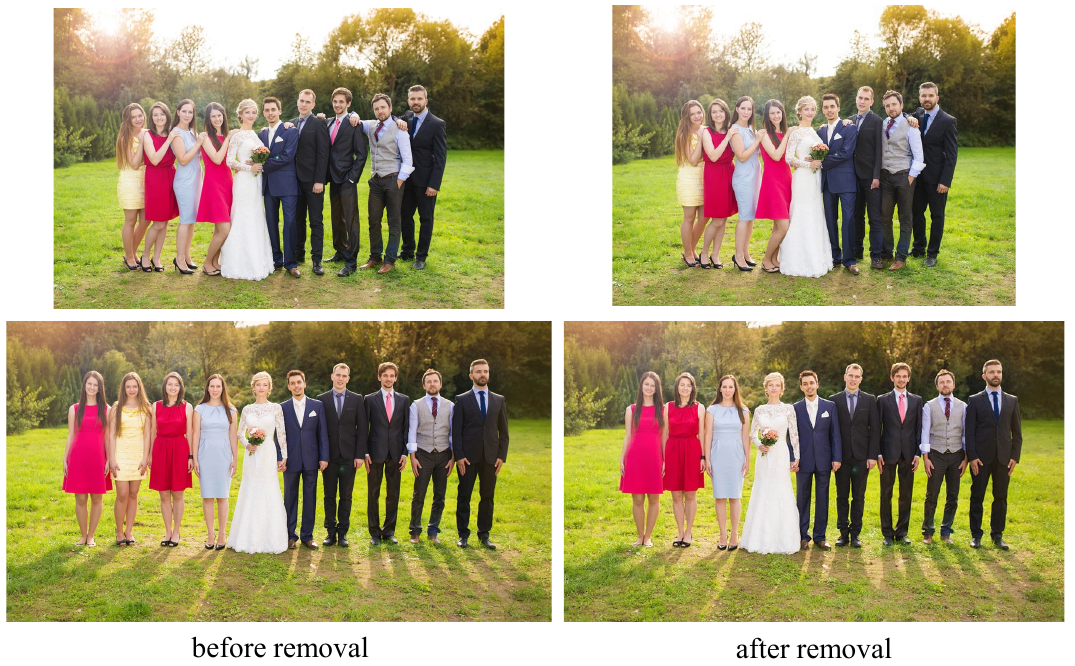}
   \end{center}
  \vspace{-20pt}
   \caption{\textbf{More Visual Results on Person Removal.}
   }
   % \vspace{-20pt}
   \label{Quali_removal}
\end{figure*}

\begin{figure*}
   \begin{center}
      \includegraphics[width=0.80\linewidth]{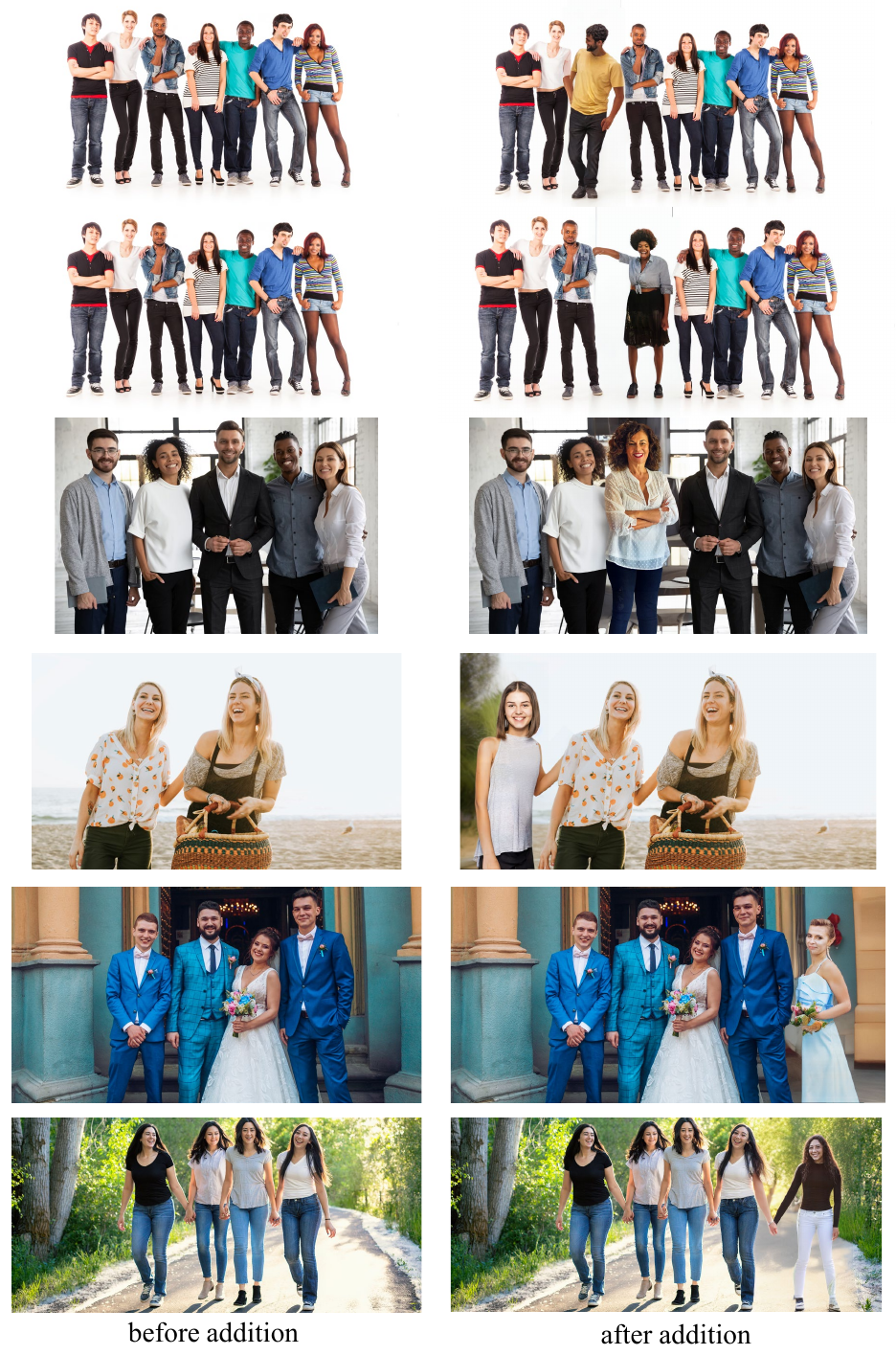}
   \end{center}
  \vspace{-20pt}
   \caption{\textbf{More Visual Results on Person Addition.}
   }
   % \vspace{-20pt}
   \label{Quali_addition}
\end{figure*}

In Fig.~\ref{Quali_editing}, Fig.~\ref{Quali_removal} and Fig.~\ref{Quali_addition}, we show more visual results generated by our proposed GroupDiff.

% \end{document}

\end{document}